\documentclass[letterpaper, 10 pt, conference]{ieeeconf}  % Comment this line out
\usepackage{FG2019}

\FGfinalcopy % *** Uncomment this line for the final submission

\IEEEoverridecommandlockouts                              % This command is only
\usepackage{diagbox}
\usepackage{makecell}
\usepackage{times}
\usepackage{epsfig}
\usepackage{graphicx}
\usepackage{amsmath}
\usepackage{sidecap}
\usepackage{amssymb}
\usepackage{graphicx}
\usepackage{subcaption}
\usepackage{multirow}
\usepackage[T1]{fontenc}
\usepackage[utf8]{inputenc}
\usepackage{textcomp}
\usepackage{siunitx}
\usepackage{tabularx,ragged2e,booktabs}
\newcolumntype{C}[1]{>{\Centering}m{#1}}

\usepackage{array}
\usepackage{color,soul}        %text color
\usepackage{algorithm}
\usepackage[noend]{algpseudocode}
\usepackage{amsmath}
\usepackage{url}
\usepackage{amsmath,amsfonts,bm}

\def\eqref#1{equation~\ref{#1}}
\def\1{\bm{1}}

\DeclareMathAlphabet{\mathsfit}{\encodingdefault}{\sfdefault}{m}{sl}
\SetMathAlphabet{\mathsfit}{bold}{\encodingdefault}{\sfdefault}{bx}{n}

\newcommand{\R}{\mathbb{R}}

\title{\LARGE \bf
Error Autocorrelation Objective Function \\ for Improved System Modeling}

\author{\parbox{16cm}{\centering
    {\large Anand Ramakrishnan$^1$, Warren Jackson $^2$, and Kent Evans$^2$}\\
    {\normalsize
    $^1$ Worcester Polytechnic Institute, Worcester, MA, USA\\
    $^2$ PARC, a Xerox Company }}
}

\begin{document}
% \IEEEoverridecommandlockouts\pubid{\makebox[\columnwidth]{978-1-7281-0089-0/19/\$31.00~\copyright{}2019 IEEE \hfill}
% \hspace{\columnsep}\makebox[\columnwidth]{ }}
% \ifFGfinal
% \thispagestyle{empty}
% \pagestyle{empty}
% \else
% \author{Anonymous submission}
% \pagestyle{plain}
% \fi

\maketitle

%%%%%%%%%%%%%%%%%%%%%%%%%%%%%%%%%%%%%%%%%%%%%%%%%%%%%%%%%%%%%%%%%%%%%%%%%%%%%%%%
\begin{abstract}
Deep learning models are trained to minimize the error between the model's output and the actual values. The typical cost function, the Mean Squared Error (MSE), arises from maximizing the log-likelihood of additive independent, identically distributed Gaussian noise.  However, minimizing MSE fails to minimize the residuals' cross-correlations, leading to over-fitting and poor extrapolation of the model outside the training set (generalization). In this paper, we introduce a "whitening" cost function, the Ljung-Box statistic, which not only minimizes the error but also minimizes the correlations between errors, ensuring that the fits enforce compatibility with an independent and identically distributed (i.i.d) gaussian noise model.  The results show significant improvement in generalization for recurrent neural networks (RNNs) (1d) and image autoencoders (2d). Specifically, we look at both temporal correlations for system-id in simulated and actual mechanical systems. We also look at spatial correlation in vision autoencoders to demonstrate that the whitening objective functions lead to much better extrapolation--a property very desirable for reliable control systems.
\end{abstract}

%%%%%%%%%%%%%%%%%%%%%%%%%%%%%%%%%%%%%%%%%%%%%%%%%%%%%%%%%%%%%%%%%%%%%%%%%%%%%%%%
\section{Introduction}
\begin{figure}[!htb]
        \centering
        \begin{subfigure}[b]{0.5\textwidth}
            \centering
            \includegraphics[trim={.0in 1.1in .0in 1.1in},clip,width=\textwidth]{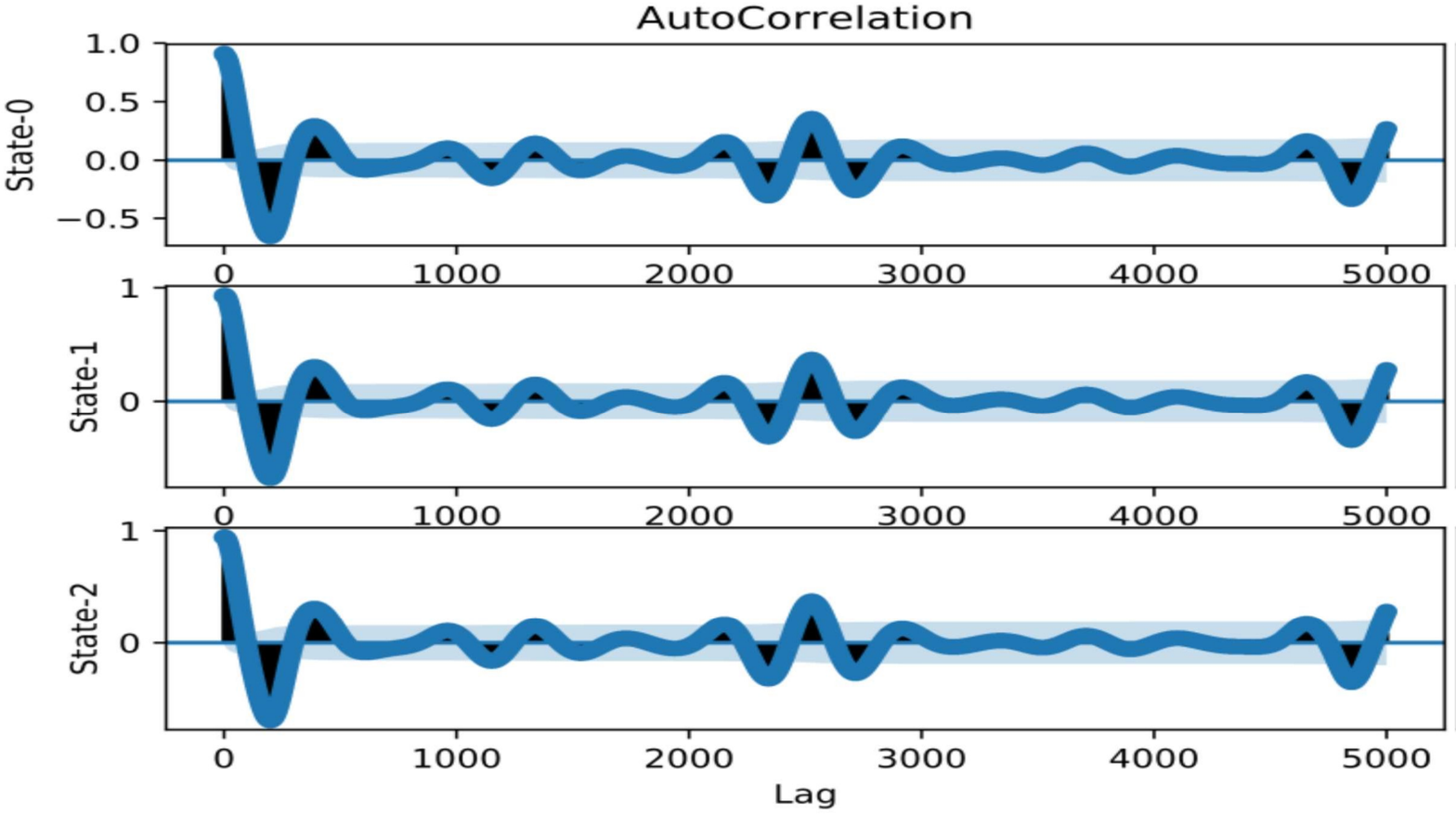}
            {{\small LSTM}}
        \end{subfigure}
        \hfill
        \begin{subfigure}[b]{0.5\textwidth}  
            \centering 
            \includegraphics[trim={.0in 1.in .0in 1.1in},clip,width=\textwidth]{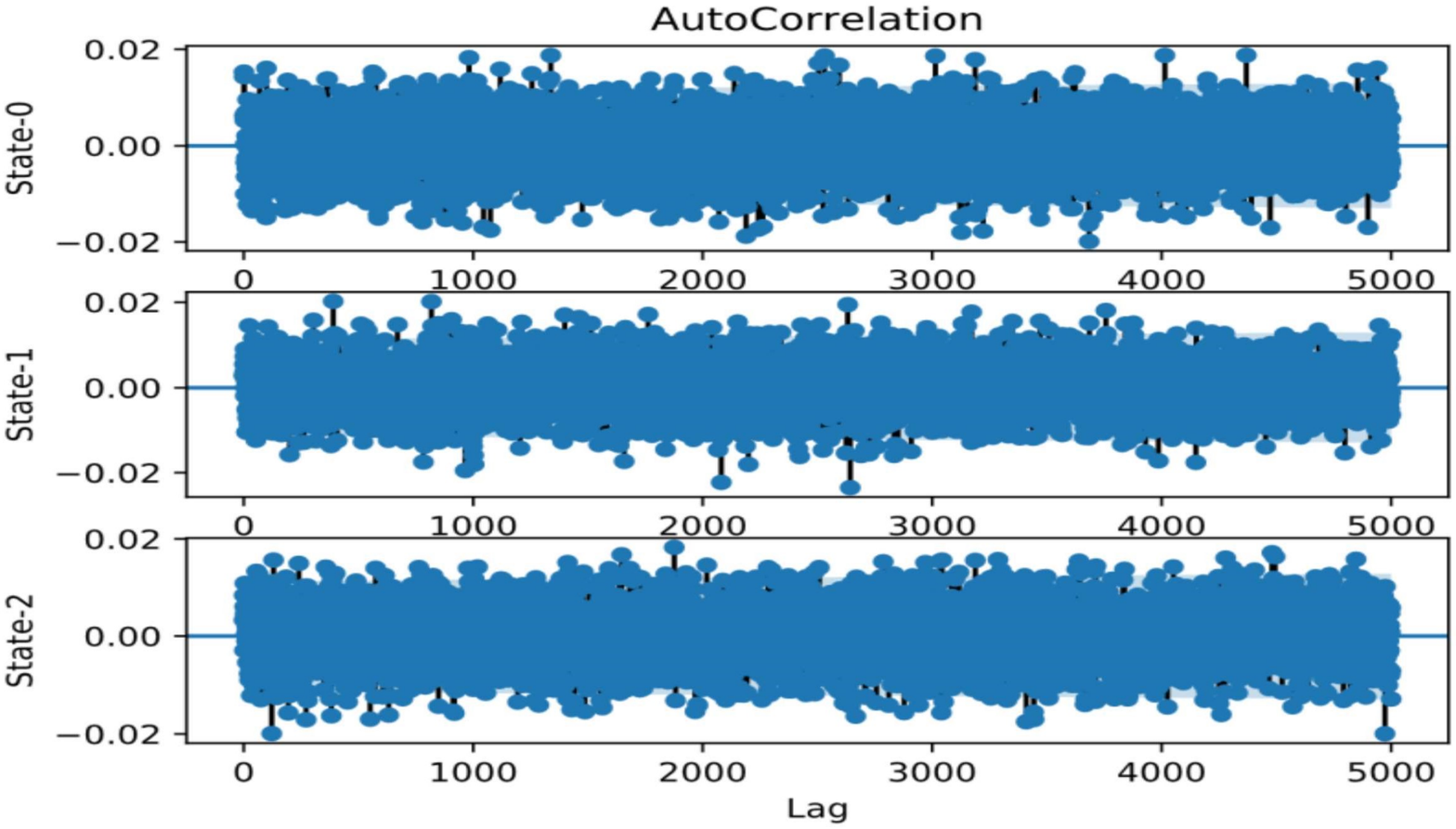}
            {{\small LSTM+LJB}}
        \end{subfigure}
        \caption[]
        {\large AutoCorrelation for an LSTM trained with an LMSE objective function and a LSTM model trained with the LMSE+ LJB cost function on the Double Pendulum.} 
        \label{fig:autocorrDP}
    \end{figure}

Most machine learning models for system identification (ID) regression problems in the temporal and spatial domains uses objective functions such as `Least Mean Squared Error' (LMSE).  These objective functions are based on the assumption that the errors are independent with no cross-correlations.  LMSE objective functions minimize the error but do not minimize the cross-correlations between errors. The presence of cross-correlations between errors leads to a myriad of problems, including exaggerated goodness of fit and inefficient model estimates \cite{rao2005error,rao2005fast}. Hence, for problems in the domain of time series prediction or image reconstruction, significant correlations between errors often remain, resulting in poor extrapolation beyond the training data. For example, consider the problem of modeling the motion of a shaft driven by a motor with the presence of backlash in the gears between them. Since large error trends dominate typical LMSE based training, small behaviors such as backlash in motors are unmodeled, unless decorrelation is enforced. Thus such a model would ultimately lead to unsatisfactory extrapolation performance. The main contributions of this paper are (1) to show that decorrelating the variables leads to greatly improved extrapolation beyond training data and (2) that decorrelating variables result in better modeling of hidden variables, i.e., small amplitude degrees of freedom masked by large ones, e.g., backlash in a motor.

\section{Whitening Cost Function}\label{sec:whiten}
In this work, minimizing the cross-correlations of the errors is applied to several robotic relevant examples, such as modeling the system-ID for the mechanical system over time. We also demonstrate the effectiveness of our method for spatial image generation. 
In particular, we use neural network models to predict $\it{n}$ steps of $\it{l}$ dimensional state vectors, $\bm{Y_t}$ $\in$ $R^{n}$ given by
\begin{equation} 
\bm{Y_t^{\it{m}}} = (y_t^{\it{m}},y_{t-1}^{\it{m}},...,y_{t-n}^{\it{m}}) \quad     m=1,...,l
\end{equation}
The model used to predict the next state vector is given by \begin{equation} \label{noiseMod}
\bm{Y_{t+1}^{\it{m}}} =  NN(\bm{Y_t^{\it{m}};W)} +\mathcal{N}(\phi) 
\end{equation}
where the neural network is trained with trainable weights $\it{W}$.  The unmodeled non-deterministic part of the data is attributed to a random  stochastic noise models $\mathcal{N}(\phi)$ with parameters $\phi$ (2rd term, Eq. (2)). The noise for each of the $\it{n}$ state vectors is assumed to be independent, identically distributed (IID).

The usual assumptions of maximum likelihood with Gaussian errors lead to LMSE. However, LMSE does not enforce error independence. To ensure independence, the cross-correlation terms of the covariance matrix must be minimized as well. In particular, the Ljung Box statistic for the whitening of uncorrelated variables can be minimized to minimize cross-correlations between residuals. 
The residuals,$\bm{R_t^m}\in\R^n$ are given by
\begin{equation}
    \bm{R_{t+1}^m}=\bm{Y_{t+1}^m} -  NN(\bm{Y_t^m;W})
\end{equation}  The covariance is given by 
\begin{equation}
 \bm{E_t<(R^m)^TR^m>}=
 \begin{bmatrix} 
    \rho_0^m & \rho_1^m &  \rho_2^m &...& \rho_{n-1}^m \\
\rho_1^m & \rho_0^m & \rho_1^m &...& \rho_{n-2}^m\\
...&...&...&...\\
\rho_{n-1}^m & \rho_{n-2}^m & \rho_{n-3}^m & ... &\rho_0 \\
\end{bmatrix}
\end{equation}
where 
\begin{equation}
        \rho_k^m= \bm{E_t<r_{t}^m r_{t-k}^m}>\quad \bm{k=(1...n-1)}
\end{equation}
are the time averaged cross-correlations.
We want to drive the cross-correlation (off-diagonal) terms to zero and minimize the diagonal terms. Minimizing the Ljung-Box statistic given by
\begin{equation}
 \mathrm{LJB(m,W)=n (n+2)\sum_{k=1}^{L}\frac{(\rho_k^m)^2}{n-k}}
\end{equation} with respect to $\bm{W}$ can help decorrelate the residuals. We propose to minimize the LSME plus the LJB term with respect to the weights, $\bm{W}$ of the NN model.  In the spatial domain involving images, the LJB objective function comprises two-dimensional cross-correlations of residuals on the image.  

\section{Related Work}
There has been work previously trying to decorrelate various aspects of neural networks. \cite{cogswell_reducing_2016} added a term to the loss function to promote decorrelation to be used in classification task.  The term is similar to the Ljung-Box statistic used in this work. Moreover, this term does not have the weightings that the Ljung-Box function does to minimize the bias. Since many of the terms included have noise, truncating the correlation coefficients is useful.

\cite{desjardins2015natural} proposed to whiten neural networks by periodically estimating the fisher matrix and imposing identity constraints on it. This led to instability while training. \cite{luo2017learning} faced a similar issue while trying to decorrelate a generalized form of the whitening matrix. In \cite{xiong_regularizing_2016}, the authors use a structured deconvolution constraint which promotes similar representations units within the same group to have strong connections and elements in different groups to utilize non-redundant structures. This required additional computational deconvolution layers.  \cite{pan_switchable_2019} and \cite{huang2018decorrelated} utilized a switchable decorrelation layer similar to batch normalization for classification invokes a layer with trainable parameters through backpropagation to facilitate the training process and decorrelate the deep layer activations. This work utilized whitening based on batches which led to poor GPU efficiency. \cite{huang2019iterative} introduced Newton iteration-based whitening for efficiency. The purpose was for classification. More recently, \cite{chen2020concept} proposed a "concept whitening layer" to create a more interpretive model by normalizing and decorrelating latent spaces. 

The methods described above apply whitening-based constraints to the model's interior structure, i.e., the weight matrices and not onto the outputs of regression. Our method involves adding an additional term to the loss function based on the output errors.

\cite{coates2011proceedings,li2015effect} studied the effects of whitening as a preprocessing step to remove pixel redundancy for training autoencoders. Works such as \cite{kobayashis2020q,asperti2020balancing} have looked at using KL-divergence to disentangle variational autoencoders to reduce correlations between latent random variables. 
In our work we look at plain autoencoders and use only whitening of errors during training.   
In addition to our method being a more simple idea, none of the above references motivated decorrelation to produce better extrapolations which are of vital importance to control systems.  

\begin{table*}[!htb]
\begin{center}
\begin{tabular}{llllllllll}
\multirow{2}{*}{\textbf{Model}} & \multicolumn{3}{c}{\textbf{State-0}} & \multicolumn{3}{c}{\textbf{State-1}} & \multicolumn{3}{c}{\textbf{State-2}} \\
                        & RMSE    & Std    & $\sum{AutoCorr}$ & RMSE   & Std    & $\sum{AutoCorr}$ & RMSE   & Std    & $\sum{AutoCorr}$ \\
Dense  & 0.02    & 0.03   & 0.51              & 0.025  & 0.04   & 0.68               & 0.01   & 0.02   & 0.25               \\
RNN & 0.0101  & 0.014  & 0.56              & 0.009  & 0.012  & 0.31               & 0.007  & 0.02   & 0.172              \\
1D-CNN & 0.009  & 0.011  & 0.41              & 0.007  & 0.010  & 0.29               & 0.007  & 0.021   & 0.191              \\
LSTM    & 0.0100  & 0.013  & 0.35           & 0.007  & 0.011  & 0.21                & 0.008 & 0.019  & 0.146             \\
Dense (LJB) & 0.04    & 0.02   & 0.029  & 0.041  & 0.03   & 0.03    & 0.03   & 0.02   & 0.005              \\
RNN (LJB)   & 0.021  & 0.019  & 0.06   & 0.011  & 0.022  &   0.058 & 0.013  & 0.033   & 0.056              \\
1D-CNN (LJB)    & 0.014  & 0.011  & {0.011}    & 0.009  & 0.019  & {0.016}  & 0.011  & 0.013  & {0.004} \\            
LSTM (LJB)    & 0.017  & 0.013  & \bf{0.009}    & 0.009  & 0.019  & \bf{0.016}  & 0.011  & 0.013  & \bf{0.004}             
\end{tabular}
\caption{Inverse Pendulum {\bf interpolating} performance of different Models for State-0, State-1 and State-2 over 10 runs. We evaluate the models using the average 10 LookForward RMS error and also present the standard deviation of the errors. We then show the $\sum$ AutoCorrelation of the residuals across output LookForward steps over 5 lags.}
\label{tab:IPinterpolating}
\end{center}
\end{table*}

\begin{table*}[!htb]
\begin{center}
\begin{tabular}{llllllllll}
\multirow{2}{*}{\textbf{Model}} & \multicolumn{3}{c}{\textbf{State-0}} & \multicolumn{3}{c}{\textbf{State-1}} & \multicolumn{3}{c}{\textbf{State-2}} \\
                        & RMSE    & Std    & $\sum{AutoCorr}$ & RMSE   & Std    & $\sum{AutoCorr}$ & RMSE   & Std    & $\sum{AutoCorr}$ \\
Dense  & 0.18    & 0.11   & 0.75              & 0.13  & 0.20   & 0.65               & 0.43   & 0.11   & 0.71               \\
RNN & 0.15  & 0.12  & 0.62              & 0.13  & 0.18  & 0.62               & 0.45  & 0.12   & 0.79              \\
1D-CNN & 0.13  & 0.11  & 0.59              & 0.11  & 0.13  & 0.65               & 0.38  & 0.15   & 0.56              \\
LSTM    & 0.12  & 0.13  & 0.41   & 0.09  & 0.11  & 0.39   & 0.29  & 0.18  & 0.49             \\
Dense (LJB) & 0.117    & 0.08   & 0.15  & 0.079  & 0.07   & 0.13    & 0.13   & 0.11   & 0.115              \\
RNN (LJB)   & 0.106  & 0.019  & 0.13   & 0.089  & 0.022  &   0.08 & 0.083  & 0.11   & 0.091              \\
1D-CNN (LJB)   & 0.081  & 0.018  & 0.09   & 0.058  & 0.020  & 0.08      & 0.079  & 0.10   & 0.088              \\
\bf{LSTM (LJB)}  & \bf{0.069}  & \bf{0.018}  & \bf{0.085}    & \bf{0.055}  & \bf{0.019}  & \bf{0.11}    & \bf{0.077}  & \bf{0.09}  & \bf{0.085}  \\           
\end{tabular}
\caption{Inverse Pendulum {\bf extrapolating} performance of different Models for State-0, State-1 and State-2. We evaluate the models using the average 10 LookForward RMS error and also present the standard deviation of the errors over 10 runs. We then show the $\sum$ AutoCorrelation of the residuals across output LookForward steps over 5 lags.}
\label{tab:IPextrapolating}
\end{center}
\end{table*}

\section{Experiment 1: Simulated System}
We perform system-ID (predicting the system behavior) for several different mechanical problems in time series prediction, an area where machine learning has not had much success \cite{makridakis2018statistical}.

In particular, problems from the OpenAI gym~\cite{brockman2016openai} such as the inverse pendulum, and a double pendulum are simulated as a function of time with Gaussian noise added. To create a system-id data set, a series of actuations are made to excite the system, and the resulting response from the simulation is collected. Training and testing data sets consist of the previous state $\bm{x(t+1)}$, the actuation $\bm{u(t)}$, and residuals $\bm{r(t)}$.  The outputs are the next time steps residual values $\bm{r(t+1)}$.  Each system's state includes various degrees of freedom, such as the angle, angular velocity, and acceleration.

The created system-ID time series includes an interpolating training set and an interpolating validation set from either restricted actuation amplitude and/or frequency. We then create an extrapolating test data set, which is used to test the models on an extrapolated activation. This dataset is created by increasing the amplitude and/or the frequency of actuation for the various models to much higher values than the interpolation training dataset. By evaluating on the extrapolation dataset we evaluate the generalization of models which is important for when the system encounters new conditions.
\subsection{Models Compared}
We utilize the most widely used set of deep models and analyze the performance improvements on them due to LJB objective function. In particular, we look at Fully Connected Networks (Dense), 1D Convolutions (1D-CNN), Recursive Neural Networks (RNN), and Long-Short Term Memory  (LSTM) on a time series task. 

\subsubsection{Neural Network Model Details}
Hyperparameter discussions are presented here. All models are trained with TensorFlow as the backend. Trained models are saved based on performance on a validation dataset of size 10000 created from the interpolation dataset. All models are trained for 1000 epochs unless specified. The initial learning rate is set at 0.01. The learning rate is halved if the validation loss plateaus for 10 consecutive epochs. Early stopping is invoked if the model's validation loss does not reduce for 30 consecutive epochs. The optimizer used is adam \cite{kingma2014adam}, and models are trained with a mini-batch of size 512. 
Batch Normalization \cite{ioffe2015batch} is used at all the layers of the model except the final layer. All trained dense Models have 5 hidden layers with 256 weights in each layer. All trained RNN and LSTM models have 5 hidden layers with 50 recurrent weight blocks in each layer. All trained 1D convolution models have 5 hidden Layers with 128 weight blocks, kernel size of 2.  

\subsection{Inverse Pendulum}
The inverted pendulum swing-up problem is a classic problem in the control literature. Model inputs are: $\cos{\Theta(t+\delta lb)}, \sin{{\Theta}(t+\delta lb)}, \dot{\Theta}(t+\delta lb), U(t+\delta lb)$ (Applied Torque) and outputs are $\cos{\Theta(t+1+\delta lf)}, \sin{{\Theta}(t+1 + \delta lf)}, \dot{\Theta}(t+1 + \delta lf)$. Here $lb$ and $lf$ are the look back window (number of previous time steps given as input to the model) and look forward window (number of forward time steps predicted by the models) which are both set to 10 time steps.

{\bf Hyperparameters:} We train our models on a training dataset of size 25000 points with control actions constrained between -0.5 and 0.5, which we call the interpolating training set. The above experiment is then repeated 10 times using 10 different random seeds.
\\{\bf Results:} The performance of different models on an interpolating test dataset (10,000 data points) is shown in Table~\ref{tab:IPinterpolating}. Here, the models' performance with just the LSME objective is slightly better than the models with the LJB+LSME objective as expected since the decorrelation adds further constraints.  The average auto-correlation of the residuals is, of course, much improved with the whitening LJB added to minimize the cross-correlations.  
We hypothesize that while models with a minimum LMSE with high residual autocorrelation perform better in the interpolation data set, they perform significantly worse on the extrapolating test data set even for the LMSE due to correlations reducing the ability to extrapolate beyond the training data set. To show this, we create an extrapolation test data set of size 25000 utilizing the entire space of control actions, i.e., between -2 and 2. The performance of the models on this dataset is shown in Table~\ref{tab:IPextrapolating}. Here we observe that the models with the whitening LJB objective do better on the LMSE as well as the autocorrelation function.  Thus, when extrapolating in actuator space or in time, it is useful to ensure that the cross-correlation is minimized.

\begin{table*}[!htb]
\begin{center}
\begin{tabular}{llllllllll}
\multirow{2}{*}{\textbf{Model}} & \multicolumn{3}{c}{\textbf{State-0}} & \multicolumn{3}{c}{\textbf{State-1}} & \multicolumn{3}{c}{\textbf{State-2}} \\
                        & RMSE    & Std    & $\sum{AutoCorr}$ & RMSE   & Std    & $\sum{AutoCorr}$ & RMSE   & Std    & $\sum{AutoCorr}$ \\
Dense & 0.009 & 0.010 & 0.77   & 0.008 & 0.011 & 0.71               & 0.012   & 0.013   & 0.55 \\
RNN &  0.006 & 0.008 & 0.61    & 0.007  & 0.009  & 0.52    & 0.007  & 0.008   & 0.49        \\
1D-CNN & 0.008 & 0.007 & 0.65   & 0.005  & 0.006  & 0.60    & 0.006  & 0.008   & 0.56       \\
LSTM & 0.005 & 0.007 & 0.39 & 0.005  & 0.005  & 0.32       & 0.006 & 0.007  & 0.32         \\
Dense (LJB) & 0.011 & 0.009 & 0.15   & 0.008  & 0.006   & 0.14     & 0.010  & 0.009   & 0.12  \\
RNN (LJB) & 0.008 & 0.006 & 0.09   & 0.009  & 0.006 &  0.07    & 0.008  & 0.006  & 0.07 \\
1D-CNN (LJB) & 0.011 & 0.009 & 0.08 & 0.010  & 0.007  & 0.08  & 0.011  & 0.009   & 0.09\\       
LSTM (LJB)    & 0.007  & 0.004  & \bf{0.01}    & 0.006  & 0.005  & \bf{0.02}  & 0.008  & 0.006  & \bf{0.02}             
\end{tabular}
\caption{Double Pendulum interpolating Performance of different Models for State-0, State-1 and State-2 over 10 runs.}
\label{tab:DPinterpolating}
\end{center}
\end{table*}

\begin{figure}
    \centering
    \includegraphics[trim={0.0in 1.0in 0.0in 1.1in},clip,width=0.45\textwidth]{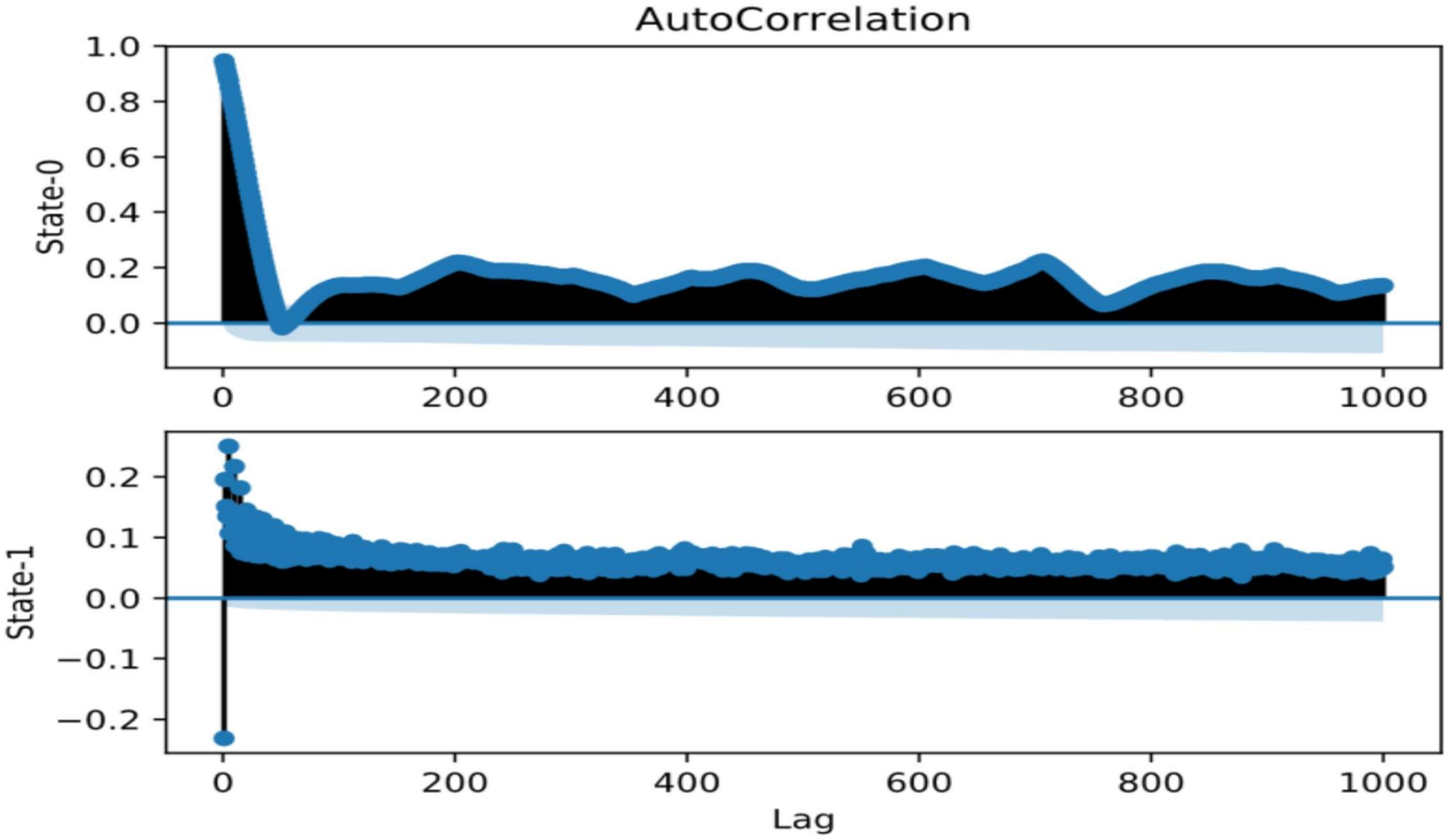}
    \includegraphics[trim={0.0in 1.0in 0.0in 1.1in},clip,width=0.45\textwidth]{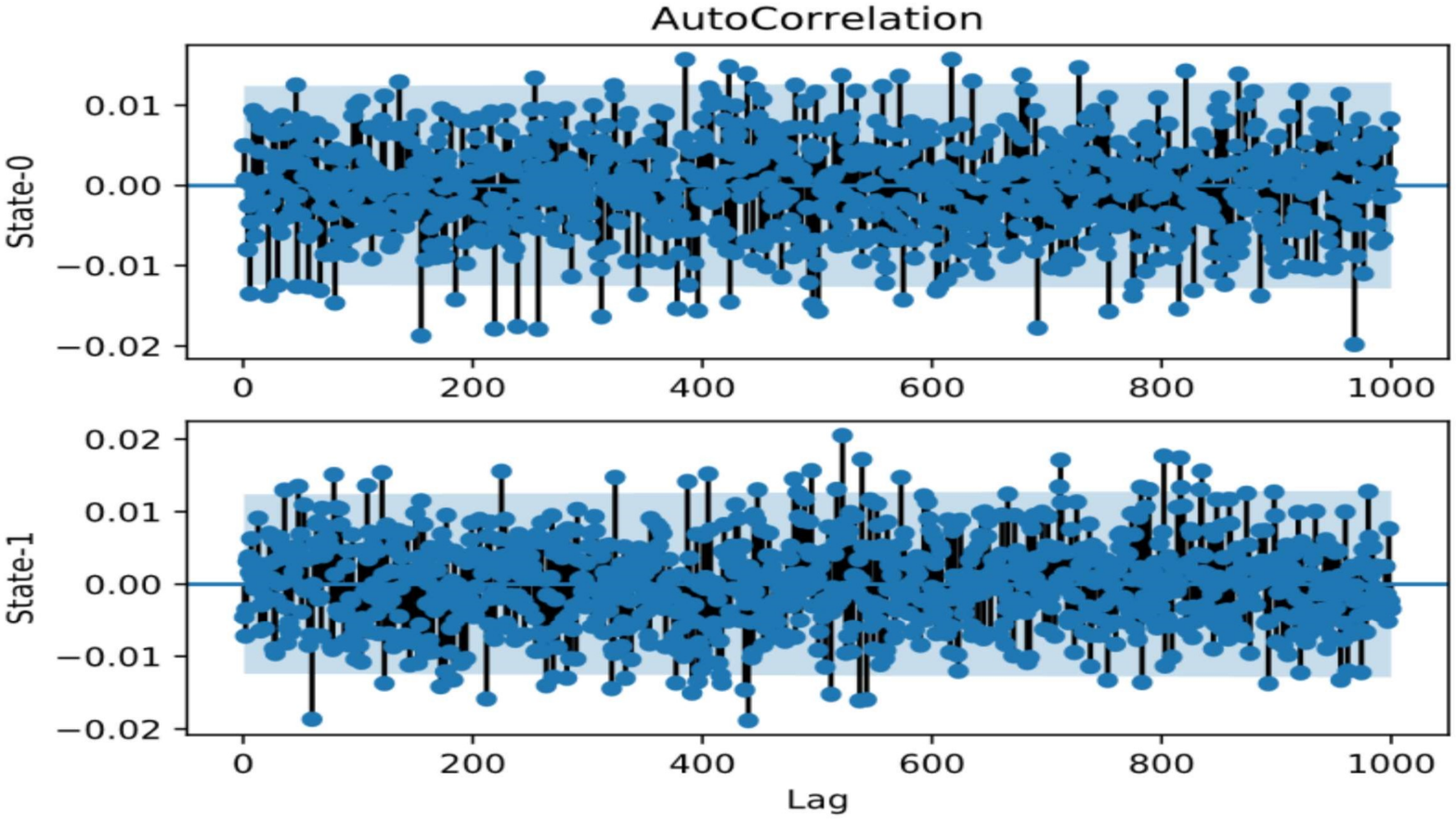}
    \caption{AutoCorrelation on the DC Motor. {\bf Top}: LSTM. {\bf Bottom}: LSTM+LJB.}
    \label{fig:autocorrDC}
\end{figure}

\subsection{Double Pendulum}
Next, we look at the double pendulum to demonstrate that the LJB function can help capture unmodeled degrees of freedom. We simulate a double pendulum in free fall under gravity. The model system consists of a large pendulum with a small second pendulum at the end starting at 90 degrees which oscillates without actuation. The input to the deep model consists only the states of the large pendulum. It should be noted that the models do not know the small pendulum's movement, which perturbs the motion of the larger pendulum. Model inputs are: ${\Theta(t)},{\dot{\Theta}(t)}, \ddot{\Theta}(t)$ and outputs are ${\Theta(t+1)}, {\dot{\Theta}(t+1)}, \ddot{\Theta}(t+1)$ of the large pendulum.

{\bf Results:} Fig.~\ref{fig:autocorrDP} shows how the LJB loss function improves the autocorrelation function of the residuals. The top part of the figure shows the autocorrelation function for the LSTM model alone.  Since the second pendulum motion is not modeled, the residual autocorrelation function exhibits the oscillation of the unmodeled pendulum's motion.  In the figure below, the LSTM model with LMSE and LJB as the loss function is shown. The autocorrelation function (with zero lag suppressed) is very close to uncorrelated. The variance of the residuals is the noise added to the simulation. This confirms the very important result that the residual decorrelation enhances the ability of a deep network to capture extrpolated structured disturbances, including unmodeled degrees of freedom. (See Table \ref{tab:DPinterpolating} for results of the various deep models on an interpolation validation set.)

\begin{table}[!htb]
\setlength\tabcolsep{3.pt}
\begin{center}
\begin{tabular}{llllllllll}
\multirow{2}{*}{\textbf{Model}} & \multicolumn{3}{c}{\textbf{State-0}} & \multicolumn{3}{c}{\textbf{State-1}}\\
& RMSE    & Std    & $\sum{AC}$ & RMSE   & Std    & $\sum{AC}$\\
Dense  & 0.18    & 0.11   & 0.75              & 0.13  & 0.20   & 0.65   \\
RNN & 0.15  & 0.12  & 0.62              & 0.13  & 0.18  & 0.62     
\\
1D-CNN & 0.13  & 0.12  & 0.55              & 0.11  & 0.14  & 0.68     
\\
LSTM    & 0.12  & 0.13  & 0.41   & 0.09  & 0.11  & 0.39   
\\
Dense (LJB) & 0.157    & 0.08   & 0.15  & 0.079  & 0.07   & 0.13    & 
\\
RNN (LJB)   & 0.146  & 0.08  & 0.13   & 0.089  & 0.05  &   0.11 & 
\\
1D-CNN (LJB)   & 0.106  & 0.07  & 0.096   & 0.071  & 0.07  &   0.11 & 
\\
\bf{LSTM (LJB)}  & {0.099}  & {0.06}  & \bf{0.085}    & {0.055}  & {0.03}  & \bf{0.09}  
\end{tabular}
\caption{DC Motor interpolating performance of different Models for State-0, State-1 and State-2. We evaluate the models using the average 10 LookForward RMS error and also present the standard deviation of the errors. We then show the $\sum$ AutoCorrelation of the residuals across output LookForward steps over 5 lags over 10 training runs.}
\label{tab:DCinterpolating}
\end{center}
\end{table}

\begin{table}[!htb]
\setlength\tabcolsep{3.pt}
\begin{center}
\begin{tabular}{llllllllll}
\multirow{2}{*}{\textbf{Model}} & \multicolumn{3}{c}{\textbf{State-0}} & \multicolumn{3}{c}{\textbf{State-1}}\\
& RMSE    & Std    & $\sum{AC}$ & RMSE   & Std    & $\sum{AC}$\\
Linear  & 3.29    & 0.11   & 0.40   & 0.09  & 0.15   & 0.24
\\
Dense  & 4.011    & 0.80   & 0.48   & 1.14  & 0.25   & 0.28
\\
RNN & 3.57  & 0.93  & 0.38  & 0.95  & 0.22  & 0.255     
\\
1D-CNN & 3.57  & 0.93  & 0.38  & 0.95  & 0.22  & 0.255     
\\
LSTM    & 3.24  & 0.88  & 0.43   & 0.88  & 0.53  & 0.19   
\\
Dense (LJB) & 3.04    & 0.18   & 0.21  & 0.21  & 0.16   & 0.15
\\
RNN (LJB)   & 2.36  & 0.15  & {0.15}   & 0.17  & 0.15  &   0.12 
\\
\bf{LSTM (LJB)}  & \bf{2.11}  & \bf{0.12}  & \bf{0.11}    & \bf{0.11}  & \bf{0.09}  & \bf{0.11}\\           
\end{tabular}
\caption{DC Motor extrapolating performance of different Models for State-0, State-1 and State-2. We evaluate the models using the average 10 LookForward RMS error and also present the standard deviation of the errors over 10 training runs. Also shown is the $\sum$ AutoCorrelation of the residuals across output LookForward steps over 5 lags over 10 training runs.}
\label{tab:DCextrapolating}
\end{center}
\end{table}

\begin{figure*}
    \includegraphics[trim={0.in .0in 0.0in 0.in},clip,width=\textwidth]{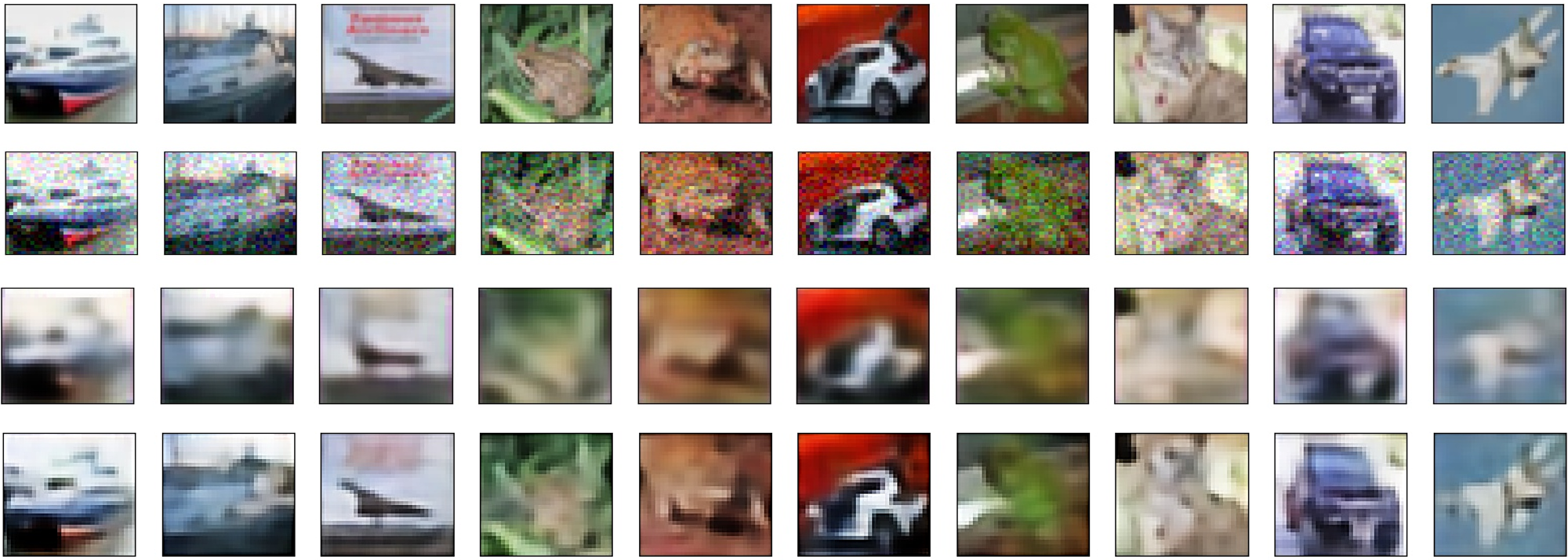}
    \caption{Reconstructed outputs from the Autoencoders for the denoising problem. Row-1 shows the true images without noise. Row-2 shows the noise added input to the autoencoder. Row 3 shows the reconstructed image using the autoencoder minimizing LMSE, and Row-4 shows the reconstructed images from the autoencoder with spatial whitening.}
    \label{fig:autoencoderdenosing}
\end{figure*}

\section{Real World System: DC Motor} \label{sec:dcmotor}
Finally, we analyze our deep models on an actual physical system. A DC motor was coupled to a rotary encoder with a 3D-printed shaft designed to purposely exhibit substantial backlash. The voltage was controlled through Pulse Width Modulation (PWM) with polarity, allowing for a range of speeds and direction changes. Position and velocity data of the shaft were collected from the encoder using a National Instruments DAQ board. For data collection, an Arduino controller was programmed to randomly vary the motor's speed and polarity and collect the data.

{\bf Hyperparameters:} We train our models on a training dataset of size 25000 points with control PWM cycles at 10HZ (interpolating training set). All models are trained using adam as the optimizer with initial learning rate of 0.001 for 1000 Epochs. The above experiment is then repeated 10 times using 10 different random seeds. The model inputs are: ${\Theta(t+\delta lb)}, \dot{\Theta}(t+\delta lb), U(t+\delta lb)$ (Applied Torque) and outputs are ${\Theta(t+1+\delta lf)}, \dot{\Theta}(t+1 + \delta lf)$. Here $lb$ and $lf$ are both set to 10 time steps.  
% The results on a interpolating validation dataset (10,000 data points) is shown in Table~\ref{tab:DCinterpolating}. We also create an extrapolation dataset of size 25000 using PWM signals up to 100HZ thus increasing the frequency by a factor of 10. The results on this dataset is shown in Table~\ref{tab:DCextrapolating}. 

{\bf Results:} Table~\ref{tab:DCinterpolating} show the performance of the deep models with and without the LJB cost function on the interpolating validation dataset. In this case, we can see that the models trained with LJB cost function perform better than their counterparts while simultaneously being less correlated than just plain LMSE training. This shows that the whitening cost function does improve the fit of the models.  
  We also create an extrapolation dataset of size 25000 using PWM signals up to 100HZ, thus increasing the frequency by a factor of 10. Table~\ref{tab:DCextrapolating} shows the performance of the deep models on this dataset. As expected, the LMSE is larger for the extrapolated data set over all the models. Importantly, our hypothesis that the models with less autocorrelation on residuals for the interpolating dataset perform better on the extrapolating dataset is validated. Thus, training with the LJB cost function has helped capture the delay of the motor's backlash, and the LMSE is thus much smaller on the extrapolating dataset. We also see that the autocorrelation values are quite small compared to the deeper models. 
(See Figure ~\ref{fig:autocorrDC} which shows the residuals' autocorrelation for the DC Motor.). 

Thus, these results demonstrate that utilizing whitening loss functions results in consistent, extrapolating models for both simulated and real systems.

\begin{table}[!htb]
\begin{center}
\begin{tabular}{llll}

\multicolumn{4}{c}{\textbf{Reconstruction Results on Cifar10}} \\ 
\textbf{Method} & \textbf{Noise} & \textbf{MSE} & \textbf{$LJB$} 
\\ 
AutoEncoder & No & 4.932E3 & 0.288 \\ 
AutoEncoder & Yes & 2.156E2 & 0.305 \\ 
AutoEncoder+LJB & No & 5.441E5 & 0.043 \\
AutoEncoder+LJB & Yes & 1.892E4 & 0.051\\ 
\end{tabular}
\caption{Reconstruction results for the autoencoders on the Cifar10 test dataset. We present the MSE and the output of the LJB cost function. }
\label{tab:Cifar}
\end{center}
\end{table}

\begin{table}[!htb]
\setlength\tabcolsep{2.5pt}
\begin{center}
\begin{tabular}{llllllllll}
\multirow{2}{*}{\textbf{Model}} & \multicolumn{3}{c}{\textbf{State-0}} & \multicolumn{3}{c}{\textbf{State-1}}\\
& RMSE    & Std    & $\sum{AC}$ & RMSE   & Std    & $\sum{AC}$\\
LSTM    & 0.12  & 0.13  & 0.41   & 0.09  & 0.11  & 0.39   
\\
LSTM (LJB)  & {0.099}  & {0.06}  & {0.085}    & {0.055}  & {0.03}  & {0.09}  
\\
Dropout    & {0.096}  & 0.06  & 0.19   & 0.053  & 0.03  & 0.11   
\\
Dropout\&LJB   & 0.095  & 0.06  & \bf{0.081}   & 0.049  & 0.02  &   \bf{0.09}  
\\
L2Norm   & 0.108  & 0.11  & 0.25   & 0.088  & 0.11  &   0.16 
\\
{L2Norm\&LJB}  & {0.116}  & {0.12}  & {0.19}    & {0.092}  & {0.12}  & {0.15}\\           
\end{tabular}
\caption{Performance of different regularizers combined with LJB function on the LSTM model for the DC Motor interpolation dataset.}
\label{tab:DCinterReg}
\end{center}
\end{table}

\begin{table}[!htb]
\setlength\tabcolsep{2.25pt}
\begin{center}
\begin{tabular}{llllllllll}
\multirow{2}{*}{\textbf{Model}} & \multicolumn{3}{c}{\textbf{State-0}} & \multicolumn{3}{c}{\textbf{State-1}}\\
& RMSE    & Std    & $\sum{AC}$ & RMSE   & Std    & $\sum{AC}$\\
LSTM    & 3.24  & 0.88  & 0.43   & 0.88  & 0.53  & 0.19   
\\
LSTM (LJB)  & 2.11  & 0.12  & 0.11 & 0.11  & 0.09  & 0.11\\           
Dropout    & 2.39  & 0.15  & 0.39   & 0.52  & 0.39  & 0.18   
\\
\bf{Dropout\&LJB}   & \bf{2.05}  & \bf{0.11}  & \bf{0.09}   & \bf{0.10}  & \bf{0.09}  &   \bf{0.10}  
\\
L2Norm   & 2.99  & 0.67  & 0.37   & 0.76  & 0.48  &   0.19
\\
{L2Norm\&LJB}  & {3.17}  & {0.69}  & {0.28}    &{0.77}  &{0.49}  & {0.18}\\           
\end{tabular}
\caption{Performance of different regularizers combined with LJB function on the LSTM model for the DC Motor extrapolation dataset.}
\label{tab:DCexterReg}
\end{center}
\end{table}

\section{Spatial Whitening in Autoencoders}
Autoencoders are models that learn an identity map for a given input space. Here we train a convolution autoencoder on the Cifar10 dataset \cite{krizhevsky2014cifar}. We perform an 80-20 split on the 50000 training images to obtain our train dataset and the validation dataset. We train our models for 100 epochs and perform early stopping with patience of 15 epochs based on the validation MSE loss. We use the best model based on the validation loss to evaluate the 10000 test images provided in the Cifar10 dataset. We compare the model's performance with plain LMSE and LMSE+LJB. We also compare the model's performance on a denoising application. We add white noise (Gaussian noise with the scaling of 0.1) to the input images and reconstruct the image without the noise.

\section{Model Details}
The architecture of the convolutional autoencoder is given below: 
Conv(64)-ReLU-Conv(64)-ReLU-BatchNorm-Maxpool-Conv(32)-ReLU-Conv(32)-ReLU-BatchNorm-Maxpool-Conv(16)-ReLU-Conv(16)-ReLU-BatchNorm-Maxpool-Conv(16)-ReLU-Conv(16)-ReLU-BatchNorm-Upsampling-Conv(32)-ReLU-Conv(32)-ReLU-BatchNorm-Upsampling-Conv(64)-ReLU-Conv(64)-ReLU-BatchNorm-Upsampling-Conv(3)-Sigmoid. 

Here the convolution layers use a kernel of size (3,3). The Maxpooling layers used a kernel of (2,2) with a stride of (2,2). The upsampling layer used a kernel of (2,2) and a stride of (2,2).

{\bf Results:} 
Table \ref{tab:Cifar} shows the model's quantitative results and figure \ref{fig:autoencoderdenosing} shows the qualitative results. As shown from the table, adding the LJB loss function improves the model's performance on the test dataset. The observed improvement is more evident for the problem of denoising autoencoders. Row-3 and Row-4 of figure \ref{fig:autoencoderdenosing}, clearly demonstrate the improved quality of the reconstructed images for reduced spatial correlations.

% \begin{figure}
%     \includegraphics[trim={0.in .0in 0.0in 0.in},clip,width=3.5in]{CifarAutoEncoderLJBvsMSE.jpg}
%     \caption{Reconstruction output from the Autoencoder+LJB model for the no noise problem. Row-1 shows the input to the autoencoder. Row-2 shows the reconstructed images from the autoencoder and Row-3 shows the error between Row-1 and Row-2}
%     \label{fig:AutoEncoderLJBvsMSE }
% \end{figure}

\section{Dropouts, L2 Norm vs LJB Function}
\label{sec:Regularizers}
Finally, we compare the LJB function with two existing regularization techniques,i.e., dropouts and L2 regularization \cite{srivastava2014dropout,krizhevsky2012imagenet}, on the model's weights. We analyze the effect of dropout and L2 regularization on the autocorrelation of the residuals. It is also imperative to verify if our proposed loss function can provide additive gains when used with existing regularization techniques to prevent overfitting. 

{\bf Procedure:} We experiment with dropouts and L2 regularization with the LSTM model on the real-world DC Motor problem defined in section \ref{sec:dcmotor}. For dropouts, we add input and recurrent dropouts to the LSTM layers. In a pilot analysis, we found that having an input dropout of 0.1 and a recurrent dropout \cite{gal2016theoretically} of 0.25 provides the best result on the interpolation validation set. We report our results using the said hyperparameters. We also perform analysis using L2-norm on all the weights layers of the LSTM model. We found that adding an L2 norm of 0.001 worked best on the interpolation validation set and thus report results based on this setting.

{\bf Results:} The results for the dropout and the L2 regularization and their combinations with the LJB function for the interpolation and extrapolation datasets on the DC Motor problem are shown in tables \ref{tab:DCinterReg}, \ref{tab:DCexterReg} respectively. As can be seen, a hybrid of dropouts and LJB loss function produces very low autocorrelation while also having the lowest LMSE on both interpolation and extrapolation datasets. It is also interesting to see that LMSE based training the LSTM model with dropouts has a moderate decorrelation effect on the residuals. Finally, we note that adding the LJB function to the L2 Norm does not provide better results. In fact, the model performs worse on the extrapolation dataset. It might be conceivable that the regularization effects of L2 Norm and LJB might be too high, thus decreasing the model's overall performance.
\section{Conclusion}

The results in our paper are important for the following three reasons. (1) Because labeled data is expensive, data efficiency is important. We can’t afford to sample the entire space of possible states; hence extrapolation is important (2) Extrapolation must be well behaved, particularly for life control situations. (3) Large amplitudes dominate the MSE, so small-amplitude behavior is not well modeled. However the correlations at different lags reflect some smaller amplitude behavior, such as the backlash. Hence decorrelation of errors helps in modeling such complex system.

Current machine learning models are not data-efficient and do not extrapolate well. Hence, models require more expensive labeled data and often provide unstable estimates of the behavior outside the data. Increases in machine learning ability to extrapolate are a major step in improving the utility of ML for real-world modeling and control.  

\bibliographystyle{ieee}
\bibliography{sigproc}  % sigproc.bib is the name of the Bibliography in this case

\begin{thebibliography}{10}\itemsep=-1pt

\bibitem{asperti2020balancing}
A.~Asperti and M.~Trentin.
\newblock Balancing reconstruction error and kullback-leibler divergence in
  variational autoencoders.
\newblock {\em IEEE Access}, 8:199440--199448, 2020.

\bibitem{brockman2016openai}
G.~Brockman, V.~Cheung, L.~Pettersson, J.~Schneider, J.~Schulman, J.~Tang, and
  W.~Zaremba.
\newblock Openai gym.
\newblock {\em arXiv preprint arXiv:1606.01540}, 2016.

\bibitem{chen2020concept}
Z.~Chen, Y.~Bei, and C.~Rudin.
\newblock Concept whitening for interpretable image recognition.
\newblock {\em Nature Machine Intelligence}, 2(12):772--782, 2020.

\bibitem{coates2011proceedings}
A.~Coates, A.~Ng, and H.~Lee.
\newblock Proceedings of the 14th international conference on artificial
  intelligence and statistics (aistats).
\newblock 2011.

\bibitem{cogswell_reducing_2016}
M.~Cogswell, F.~Ahmed, R.~Girshick, L.~Zitnick, and D.~Batra.
\newblock Reducing {Overfitting} in {Deep} {Networks} by {Decorrelating}
  {Representations}.
\newblock {\em arXiv:1511.06068 [cs, stat]}, June 2016.
\newblock arXiv: 1511.06068.

\bibitem{desjardins2015natural}
G.~Desjardins, K.~Simonyan, R.~Pascanu, and K.~Kavukcuoglu.
\newblock Natural neural networks.
\newblock {\em arXiv preprint arXiv:1507.00210}, 2015.

\bibitem{gal2016theoretically}
Y.~Gal and Z.~Ghahramani.
\newblock A theoretically grounded application of dropout in recurrent neural
  networks.
\newblock In {\em Advances in neural information processing systems}, pages
  1019--1027, 2016.

\bibitem{huang2018decorrelated}
L.~Huang, D.~Yang, B.~Lang, and J.~Deng.
\newblock Decorrelated batch normalization.
\newblock In {\em Proceedings of the IEEE Conference on Computer Vision and
  Pattern Recognition}, pages 791--800, 2018.

\bibitem{huang2019iterative}
L.~Huang, Y.~Zhou, F.~Zhu, L.~Liu, and L.~Shao.
\newblock Iterative normalization: Beyond standardization towards efficient
  whitening.
\newblock In {\em Proceedings of the IEEE/CVF Conference on Computer Vision and
  Pattern Recognition}, pages 4874--4883, 2019.

\bibitem{ioffe2015batch}
S.~Ioffe and C.~Szegedy.
\newblock Batch normalization: Accelerating deep network training by reducing
  internal covariate shift.
\newblock {\em arXiv preprint arXiv:1502.03167}, 2015.

\bibitem{kingma2014adam}
D.~P. Kingma and J.~Ba.
\newblock Adam: A method for stochastic optimization.
\newblock {\em arXiv preprint arXiv:1412.6980}, 2014.

\bibitem{kobayashis2020q}
T.~Kobayashis.
\newblock q-vae for disentangled representation learning and latent dynamical
  systems.
\newblock {\em IEEE Robotics and Automation Letters}, 5(4):5669--5676, 2020.

\bibitem{krizhevsky2014cifar}
A.~Krizhevsky, V.~Nair, and G.~Hinton.
\newblock The cifar-10 dataset.
\newblock {\em online: http://www. cs. toronto. edu/kriz/cifar. html}, 55,
  2014.

\bibitem{krizhevsky2012imagenet}
A.~Krizhevsky, I.~Sutskever, and G.~E. Hinton.
\newblock Imagenet classification with deep convolutional neural networks.
\newblock {\em Advances in neural information processing systems},
  25:1097--1105, 2012.

\bibitem{li2015effect}
Z.~Li, Y.~Fan, and W.~Liu.
\newblock The effect of whitening transformation on pooling operations in
  convolutional autoencoders.
\newblock {\em EURASIP Journal on Advances in Signal Processing},
  2015(1):1--11, 2015.

\bibitem{luo2017learning}
P.~Luo.
\newblock Learning deep architectures via generalized whitened neural networks.
\newblock In {\em International Conference on Machine Learning}, pages
  2238--2246. PMLR, 2017.

\bibitem{makridakis2018statistical}
S.~Makridakis, E.~Spiliotis, and V.~Assimakopoulos.
\newblock Statistical and machine learning forecasting methods: Concerns and
  ways forward.
\newblock {\em PloS one}, 13(3):e0194889, 2018.

\bibitem{pan_switchable_2019}
X.~Pan, X.~Zhan, J.~Shi, X.~Tang, and P.~Luo.
\newblock Switchable {Whitening} for {Deep} {Representation} {Learning}.
\newblock In {\em 2019 {IEEE}/{CVF} {International} {Conference} on {Computer}
  {Vision} ({ICCV})}, pages 1863--1871, Seoul, Korea (South), Oct. 2019. IEEE.

\bibitem{rao2005error}
Y.~N. Rao, D.~Erdogmus, and J.~C. Principe.
\newblock Error whitening criterion for adaptive filtering: theory and
  algorithms.
\newblock {\em IEEE Transactions on signal processing}, 53(3):1057--1069, 2005.

\bibitem{rao2005fast}
Y.~N. Rao, D.~Erdogmus, G.~Y. Rao, and J.~C. Principe.
\newblock Fast error whitening algorithms for system identification and control
  with noisy data.
\newblock {\em Neurocomputing}, 69(1-3):158--181, 2005.

\bibitem{srivastava2014dropout}
N.~Srivastava, G.~Hinton, A.~Krizhevsky, I.~Sutskever, and R.~Salakhutdinov.
\newblock Dropout: a simple way to prevent neural networks from overfitting.
\newblock {\em The journal of machine learning research}, 15(1):1929--1958,
  2014.

\bibitem{xiong_regularizing_2016}
W.~Xiong, B.~Du, L.~Zhang, R.~Hu, and D.~Tao.
\newblock Regularizing {Deep} {Convolutional} {Neural} {Networks} with a
  {Structured} {Decorrelation} {Constraint}.
\newblock In {\em 2016 {IEEE} 16th {International} {Conference} on {Data}
  {Mining} ({ICDM})}, pages 519--528, Dec. 2016.
\newblock ISSN: 2374-8486.

\end{thebibliography}
% ACM needs 'a single self-contained file'!
%

\end{document}